\documentclass[]{interact}
\usepackage[dvipsnames]{xcolor}
\usepackage{epstopdf}
\usepackage[caption=false]{subfig}

\usepackage[longnamesfirst,sort]{natbib}

\theoremstyle{plain}

\theoremstyle{definition}

\theoremstyle{remark}

\usepackage{longtable}
\usepackage{lipsum}
\usepackage{comment}
\usepackage{graphicx}
\usepackage[thinlines]{easytable}
\usepackage{url}
\usepackage{tabulary}
\usepackage{array}
\usepackage{array}
\newcolumntype{L}[1]{>{\raggedright\let\newline\\\arraybackslash\hspace{0pt}}m{#1}}
\newcolumntype{C}[1]{>{\centering\let\newline\\\arraybackslash\hspace{0pt}}m{#1}}
\newcolumntype{R}[1]{>{\raggedleft\let\newline\\\arraybackslash\hspace{0pt}}m{#1}}
\newcolumntype{L}[1]{>{\raggedright\let\newline\\\arraybackslash\hspace{0pt}}m{#1}}
\newcolumntype{C}[1]{>{\centering\let\newline\\\arraybackslash\hspace{0pt}}m{#1}}
\newcolumntype{R}[1]{>{\raggedleft\let\newline\\\arraybackslash\hspace{0pt}}m{#1}}
\usepackage{ragged2e}
\newcolumntype{P}[1]{>{\RaggedRight\hspace{0pt}}p{#1}}
\usepackage{longtable}
\usepackage{lipsum}
\usepackage[utf8]{inputenc}

\begin{document}

\articletype{REVIEW ARTICLE}

\title{Recent Advancements in Machine Learning For Cybercrime Prediction}

\author{
\name{Lavanya Elluri\textsuperscript{*a},
Varun Mandalapu\textsuperscript{*b},
Piyush Vyas\textsuperscript{*a},
\thanks{CONTACT: Varun Mandalapu Email:varunm1@umbc.edu, \textsuperscript{*}These authors contributed equally to this work.}
and Nirmalya Roy\textsuperscript{b}}
\affil{\textsuperscript{a}{Subhani Department of Computer Information Systems, Texas A\&M University - Central Texas, 1001 Leadership Pl, Killeen, 76549, TX, USA}, 
\textsuperscript{b}{Information Systems, University of Maryland Baltimore County, 1000 Hilltop Cir, Baltimore, 21250, MD, USA}
}}

\maketitle

\begin{abstract}
Cybercrime is a growing threat to organizations and individuals worldwide, with criminals using sophisticated techniques to breach security systems and steal sensitive data. This paper aims to comprehensively survey the latest advancements in cybercrime prediction, highlighting the relevant research. For this purpose, we reviewed more than 150 research articles and discussed 50 most recent and appropriate ones. We start the review with some standard methods cybercriminals use and then focus on the latest machine and deep learning techniques, which detect anomalous behavior and identify potential threats. We also discuss transfer learning, which allows models trained on one dataset to be adapted for use on another dataset. We then focus on active and reinforcement learning as part of early-stage algorithmic research in cybercrime prediction. Finally, we discuss critical innovations, research gaps, and future research opportunities in Cybercrime prediction. This paper presents a holistic view of cutting-edge developments and publicly available datasets.
\end{abstract}


\section{Introduction}\label{sec1}

Cyberattacks' increasing frequency and complexity have made cybersecurity a top priority for governments, businesses, and individuals. Cybercrime has significantly threatened digital assets' confidentiality, integrity, and availability, causing financial losses, reputational damages, and even physical harm. According to a notice by Cybersecurity Ventures, global cybercrime damages are expected to reach $\$10.5$ trillion annually by 2023 \cite{morgan2022top5}, up from $\$3$ trillion in 2015, making it the fastest-growing crime in the world with cyberattacks occurring every 11 seconds. In 2020, Federal Bureau of Investigation (FBI) reported a 400 $\%$ increase in cybercrime incidents compared to pre-pandemic levels \cite{smith2020fbi}, and a report by McAfee \cite{gann2020hidden} estimated that the global cost of cybercrime has increased by 50$\%$ since 2018 due to the COVID-19 pandemic. According to a survey conducted by International Business Machines (IBM) \cite{ibm2022cost}, the average cost of a data breach in 2022 was $\$9.44$ million, and it took an average of 243 days to identify and 84 days to contain a breach. Therefore, there is an urgent need for innovative and effective approaches to predict and prevent cybercrime.

Machine learning (ML), deep learning (DL), and transfer learning (TL) are emerging technologies that have shown tremendous potential in cybersecurity applications. These techniques allow the analysis of massive amounts of data, which can be used to identify patterns and anomalies that indicate potential cyber threats \cite{apruzzese2018effectiveness}. ML algorithms can learn from historical data to develop models to predict future events. DL algorithms use neural networks to learn and represent data in multiple layers, enabling more complex and accurate predictions. TL leverages pre-trained models to improve the performance of new models on related tasks. The use of these ML, DL, and TL in predicting cybercrime has gained significant attention in recent years \cite{dilek2015applications}. Cybercrime prediction can enable proactive measures to mitigate risks and prevent attacks before they occur. The ability to predict cyber threats is particularly crucial for critical infrastructures, such as energy, transportation, and healthcare, which are essential for the functioning of society. In addition, cybercrime prediction can assist law enforcement agencies in identifying and apprehending cybercriminals \cite{perry2013predictive}.

ML, DL, and TL have been applied in various cybersecurity domains to predict and prevent cybercrime. For example, Anomaly detection using ML algorithms is a popular technique for identifying unauthorized access to sensitive data \cite{chayal2021review, sinaeepourfard2019cybersecurity}. In this method, ML algorithms are trained on a dataset of normal network activity to learn patterns of normal behavior. When the algorithm detects deviations from these patterns, it flags the activity as anomalous and raises an alert. For instance, Support Vector Machines (SVMs) and Random Forests (RFs) are two popular ML algorithms that are used in intrusion detection systems to identify suspicious behavior in network traffic \cite{shams2018novel,resende2018survey}. Similarly, DL algorithms have been used to identify and classify malware and phishing attacks \cite{saha2020phishing,aljabri2022phishing}. DL models are particularly effective at identifying previously unseen malware, as they can detect patterns and features that traditional signature-based detection systems may miss. For instance, Convolutional Neural Networks (CNNs) and Recurrent Neural Networks (RNNs) have been used to analyze the behavior of malware and identify common features that can be used to detect new strains \cite{chen2019applying,jha2020recurrent}.
Similarly, Natural Language Processing (NLP) techniques have been applied to analyze the language used in phishing emails and identify common patterns that can be used to identify new phishing attacks \cite{salloum2021phishing, egozi2018phishing}. Finally, TL has been used in various cybersecurity applications, including intrusion detection and vulnerability assessment \cite{weiss2017detection}. In these cases, pre-trained models are leveraged to improve the accuracy of predictions. For example, a pre-trained model trained on an extensive network traffic dataset can be used to identify suspicious activity in a smaller, more targeted dataset. Similarly, a model trained on vulnerability assessments in one field can be transferred to another to improve the accuracy of predictions in that domain. By using pre-trained models \cite{han2019recognizing}, cybersecurity professionals can enhance the accuracy of their predictions and lower the time and effort required to train models from scratch.

ML, DL and TL are widely employed in cybersecurity, particularly in predicting cyber-physical system (CPS) attacks like those in smart grids and autonomous vehicles \cite{al2018cyber,sedjelmaci2020cyber}. The reliability of CPSs is vital, and ML, DL, and TL help analyze their data to preempt threats. Network security benefits from these techniques too, especially in thwarting distributed denial of service (DDoS) attacks \cite{sachdeva2022machine,dasgupta2022machine}. ML detects DDoS attacks by scrutinizing network traffic patterns, while DL employs neural networks to classify traffic as normal or malicious \cite{guo2022gld}. TL augments DDoS detection by leveraging pre-trained models for feature extraction. Additionally, Active Learning (AL) and Reinforcement Learning (RL) based approaches \cite{wang2023implementing,yin2020apply} are emerging to address earlier limitations in cybercrime prediction, selecting informative samples and refining model performance. These promising techniques are at an early research stage, potentially enhancing prediction models significantly as research in active and reinforcement learning in cybercrime prediction progresses.

\begin{figure*}[htbp]%
    \begin{center}
        \includegraphics[scale=0.45]{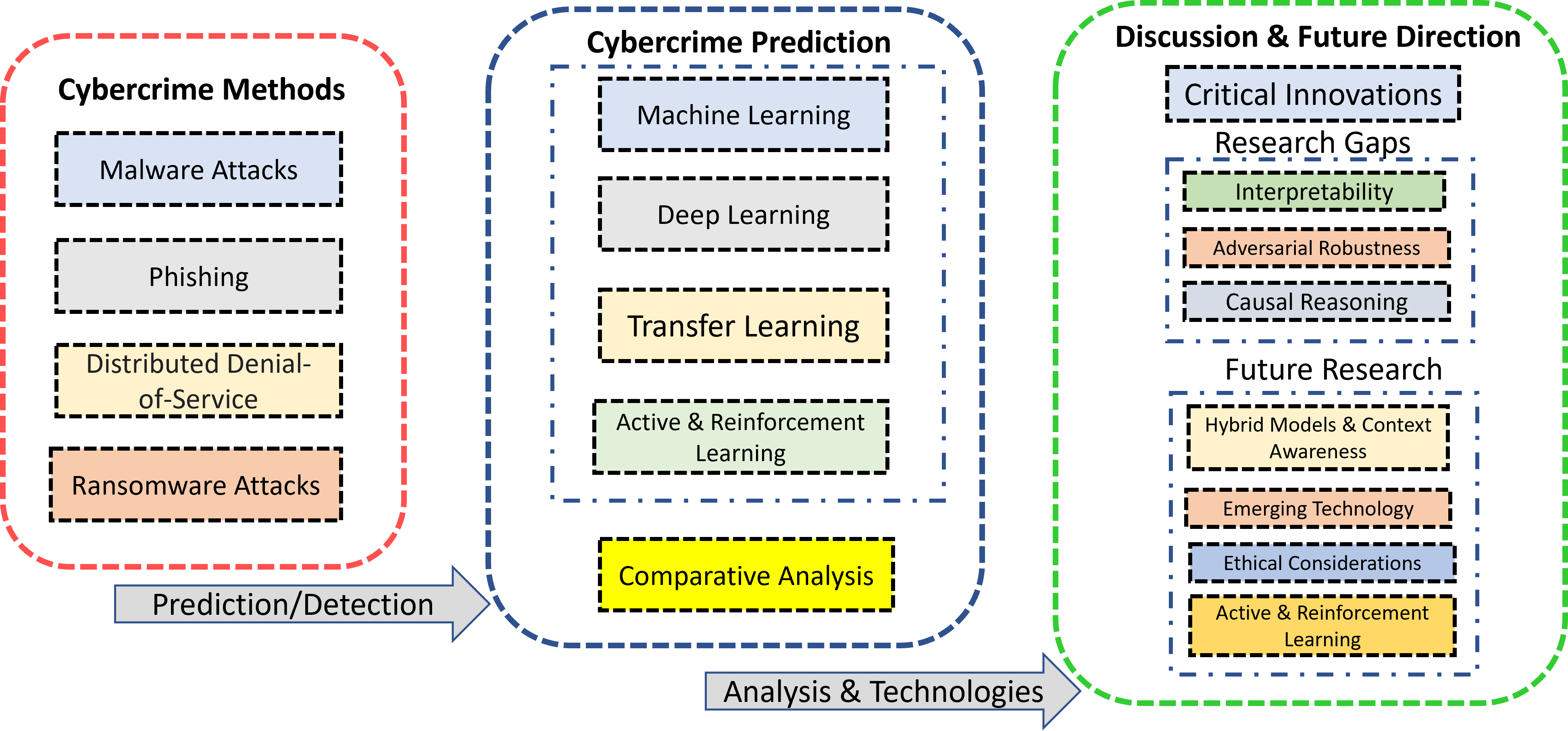}
        \caption{\centering{Overview of Cybercrime Review: Cybercrime Methods, Prediction, and Discussion}}
        \label{fig:CC_taxonomies}
    \end{center}
\end{figure*}

This survey research article aims to provide an overview of the state-of-art techniques in cybercrime prediction using ML, DL, and TL. We also discuss early-stage methodologies like AL and RL emerging in cybercrime. Figure \ref{fig:CC_taxonomies} details the scope of taxonomies in this study. 

\section{Research Methodology}\label{sec:RM}

This study consolidates effective algorithms for predicting cybercrime inspired by the field's focus and progress. Expanding to
encompass machine learning, deep learning, transfer learning, and adaptive
learning techniques, this investigation scrutinizes relevant literature from 2018 -
2023, sourced from multiple databases.
 
This review included predominantly used terms across the selected papers, employing the wildcard character ``*" to cover possible term alternatives in the  Institute of Electrical and Electronics Engineers (IEEE), Science Direct, and Association for Computing Machinery (ACM) databases. Following are the search queries.
 
IEEE  query: 

((``Document Title": ``e-crime" OR ``Document Title": "cyber*crime") AND (``Document Title": ``predic*" OR ``Document Title": ``detec*" OR ``Document Title": ``recogni*" OR ``Document Title": ``machine learning" OR ``Document Title":``deep learning" OR ``Document Title":``transfer learning" OR ``Document Title":``nlp" OR ``Document Title":``natural language processing")))

Science Direct Query:

(``e-crime" OR ``cybercrime") AND ( ``prediction" OR ``detection" OR ``recognition" OR ``machine learning" OR ``deep learning" OR ``transfer learning" OR ``nlp" OR ``natural language processing")

\begin{figure}
\begin{center}
\includegraphics[scale=0.55]{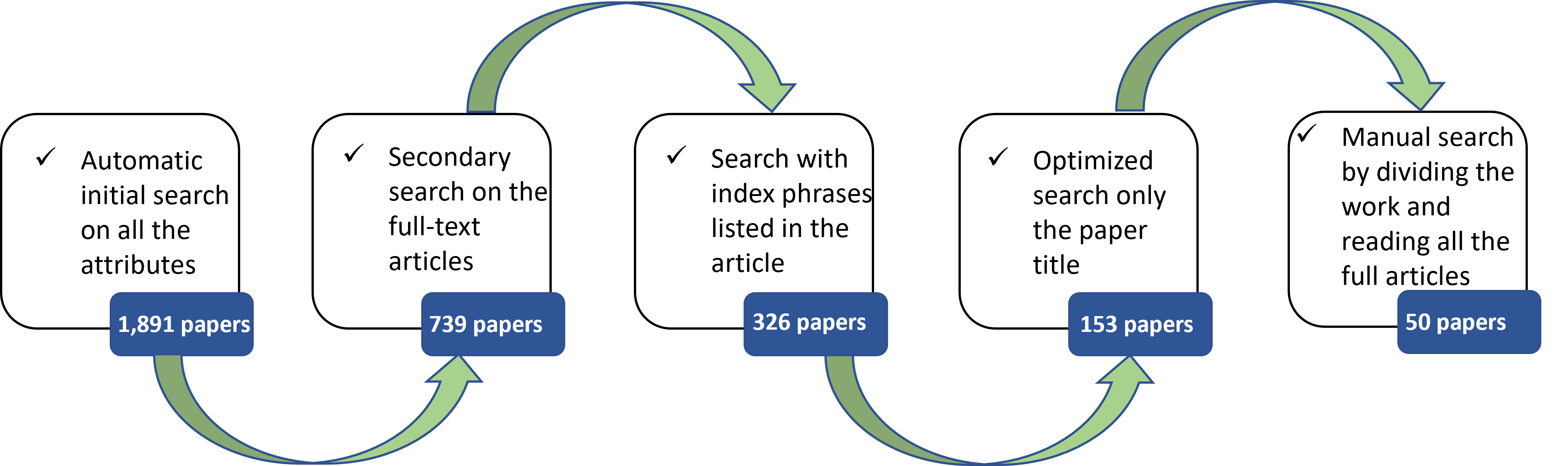}
\caption{\centering{Procedure for Cybercrime Prediction Literature Extraction}}
\label{fig:steps}
\end{center}
\end{figure}

ACM Query:

[[Title: ``cyber*crime"] OR  [Title: ``e*crime"]]AND [[Title: ``predic*"] OR [Title: ``detec*"] OR [Title: ``recogni*"] OR [Title: ``machine learning"] OR [Title: ``deep learning"] OR [Title: ``transfer learning"] OR [Title: ``nlp"] OR [Title: ``natural language processing"]]

Our study's primary aims include a comprehensive review of utilized algorithms and assisting the research community by pinpointing relevant datasets. Unrelated works were excluded via meticulous application of database search filters, employing a blend of automated and manual search methodologies as illustrated in figure \ref{fig:steps}.

Adherence to PRISMA methodology \cite{moher2009preferred} was crucial in maintaining rigor in this systematic literature review (SLR). Our search process included key term identification, individual database syntax query construction, and a focus on distinct digital research libraries to eliminate duplications.

To mitigate potential biases, which are a vital concern in accordance with PRISMA standards, all steps of the review process were conducted by multiple researchers independently where possible, and consensus was reached in all decisions. Additionally, the inclusion and exclusion criteria were defined explicitly and applied uniformly across all the databases, ensuring a consistent approach to paper selection.

The initial search, despite stringent filters on metadata and full-text papers, yielded 739 papers. A further filter applied to document titles reduced this to 153. A thorough examination of titles, keywords, and abstracts followed, resulting in irrelevant articles being discarded. Adhering to PRISMA guidelines, we selected 50 papers based on inclusion and exclusion criteria reflecting their relevance to the subject matter and a focus on state-of-the-art techniques. Inclusion Criteria - All previous research that examined cybercrimes in general or different cybercrime tactics have been included. We also focused on the publications' use of cutting-edge technologies like machine, deep, transfer, and adaptive learning. In addition, regardless of the article's qualitative or quantitative behavior, we chose those whose focus or aims aligned with the cybercrime prediction. Exclusion criteria - All publications that conducted literature reviews and meta-analyses were disregarded. Additionally, we have eliminated any publications that did not specifically mention cybercrime prediction methods in the context of state-of-the-art techniques.

\section{Cybercrime Methods}
\label{sec:CC_Methods}
\begin{figure*}[htbp!]
    \begin{center}
        \includegraphics[scale=0.5]{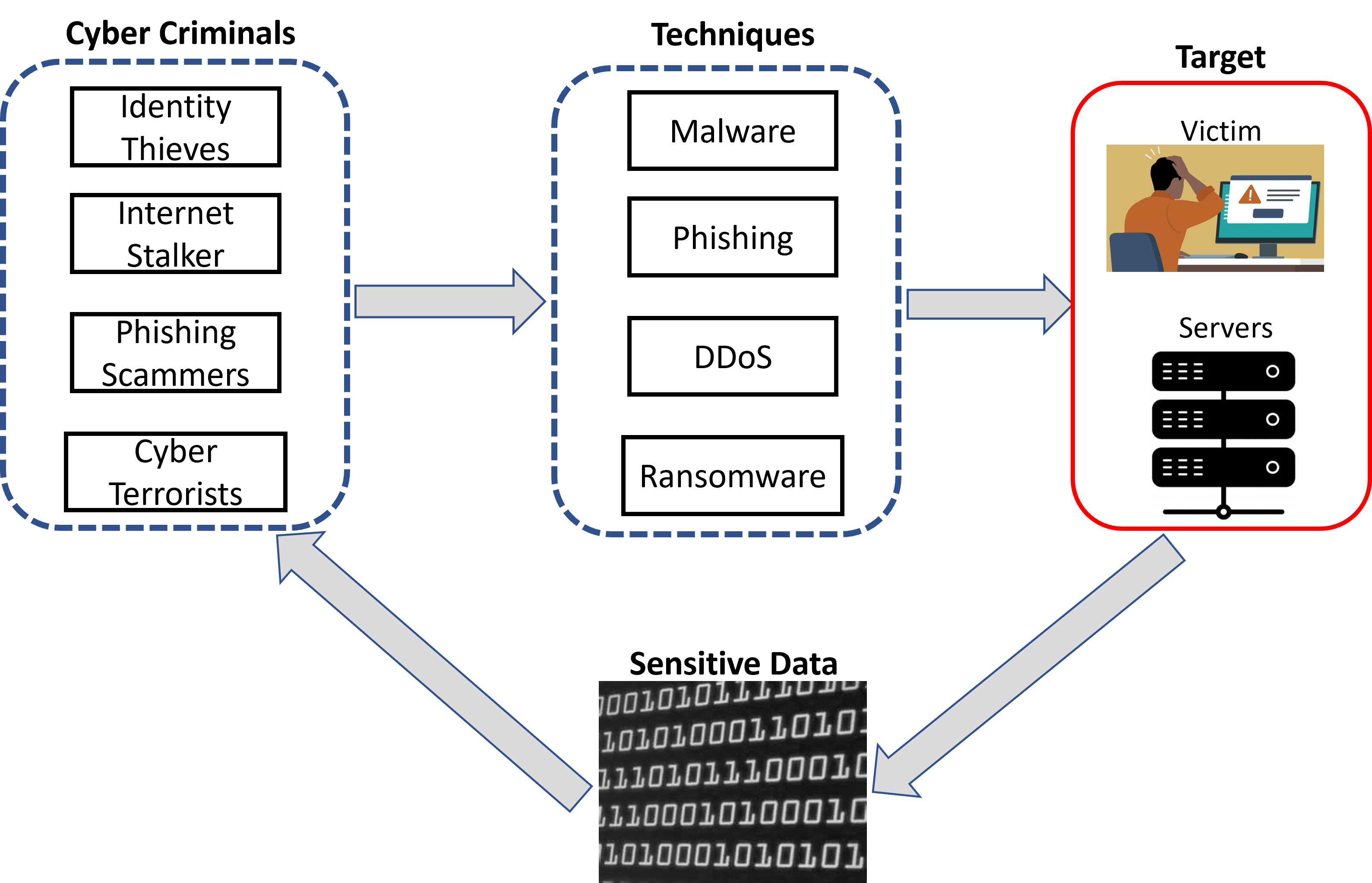}
        \caption{\centering{Components of Cybercrime: Cyber Criminals, Techniques and Target}}
        \label{fig:CC_overview}
    \end{center}
\end{figure*}
Cybercrime is a criminal activity that takes place in the digital world. Cybercriminals use various methods to perform cybercrimes and cause damage to individuals and organizations, as shown in figure \ref{fig:CC_overview}. Below, we thoroughly discuss some of the most common methods used to perform cybercrime. he below primary methods were selected based on the latest research by \cite{aslan2023comprehensive} informing how these attacks are increasing in complexity while the knowledge needed to develop these attacks is easily accessible and attainable. It is worth noting that there are numerous emerging techniques in the realm of cyberattacks. However, to maintain the focus and scope of our work, we have specifically included the below methods due to their significance and complexity within the broader landscape of cybercrime.

\textbf{Malware Attacks}:  Malware attacks are a standard method used by cybercriminals to gain unauthorized access to computer systems and networks \cite{furnell2019cyber,alenezi2020evolution}. Malware is software that is designed to disable or damage computer systems. Cybercriminals can use malware attacks to access sensitive information or disable a computer network \cite{alenezi2020evolution}. Malware attacks can be performed through various methods, such as email attachments, malicious websites, or vulnerabilities in software \cite{broadhead2018contemporary} for stealing sensitive information, encrypting files, or launching a distributed denial-of-service (DDoS) attack. Recent research such as \cite{saad2019curious} has focused on developing new methods to detect and prevent malware attacks.

\textbf{Phishing}: Phishing is a type of cybercrime that involves using social engineering techniques to mislead people into disclosing sensitive information, such as usernames, passwords, and credit card numbers \cite{alkhalil2021phishing}. Phishing attacks typically involve sending messages or emails that seem to be from a legitimate source, such as an online retailer or bank, to persuade the recipient to click on a link or open an attachment that installs malware or redirects to a fake website. Researchers such as \cite{alhogail2021applying} are developing new methods to detect and prevent these attacks by using ML, DL, and NLP algorithms.

\begin{figure*}[htbp!]
    \begin{center}
        \includegraphics[scale=0.5]{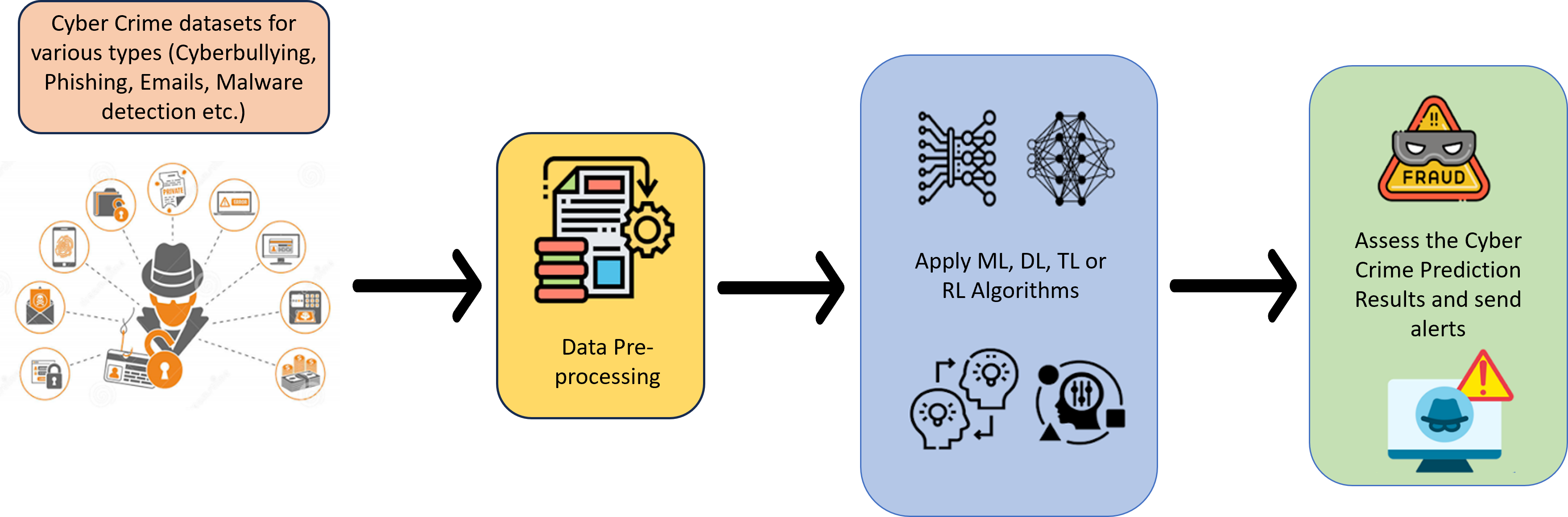}
        \caption{\centering{Steps involved in typical cybercrime prediction}}
        \label{fig:overview}
    \end{center}
\end{figure*}
\textbf{DDoS}: Distributed denial-of-service (DDoS) attacks are a type of cyberattack that seeks to disrupt the availability of a targeted website or service by overwhelming it with traffic from multiple sources \cite{al2021ddos}. DDoS attacks can be performed through various methods, such as botnets, amplification attacks, and application-layer attacks. Botnets are networks of compromised computers retained by cybercriminals to perform DDoS attacks. Amplification attacks involve sending a small request to a server that generates a much larger response toward the victim \cite{griffioen2021scan}. Application-layer attacks target the application layer of a website or service, such as by sending many requests designed to overload the application. Recent research has focused on developing detection techniques based on ML, DL, and TL approaches \cite{mittal2022deep,yusof2019systematic}.

\textbf{Social Engineering}: Social engineering attacks are a type of cybercrime involving manipulating people to reveal sensitive information or perform actions against their interests \cite{salahdine2019social}. Social engineering attacks are performed through various methods, such as phishing emails, pretexting, baiting, or quid pro quo. These attacks are becoming more sophisticated and harder to detect as they rely on exploiting human psychology and emotions, making them difficult to detect using standard security measures such as firewalls or antivirus software. Recent research such as \cite{pradeepa2022malicious} has focused on developing new methods to detect and prevent social engineering attacks.

\textbf{Ransomware Attacks}: Ransomware attacks involve encrypting the victim's data and demanding a ransom payment in exchange for the decryption key \cite{reshmi2021information}. These attacks typically involve gaining access to the victim's computer system through various methods, such as phishing emails, software vulnerabilities, or remote desktop connections. Once the attacker acquires access to the victim's system, they use specialized malware to encrypt the victim's files and demand payment in trade for the decryption key. The attackers typically demand payment in cryptocurrency, such as Bitcoin, to make it difficult to trace the transaction \cite{mos2020growing}. Ransomware attacks are detected by analyzing network traffic, behavioral patterns, and system log datasets \cite{chen2019automated}. When an attack is commencedMarkov model-based algorithms showed promising results in detecting these attacks as multiple states occur  \cite{hwang2020two}.

As the methods used by cybercriminals are becoming increasingly hard to detect, researchers are developing new methods to detect and prevent these attacks, focusing on ML, DL, and TL algorithms. The following sections will show that researchers are constantly developing new and innovative methods to enhance the effectiveness of these algorithms.

\section{Cybercrime Prediction}\label{sec:CC_Predict}

The rise of cybercrime is a growing concern in today's digital world, and researchers focused on developing effective methods to predict and prevent these threats. ML, DL, TL, and adaptive learning have emerged as powerful tools in the fight against cybercrime. These techniques leverage algorithms and neural networks to analyze large datasets and identify patterns and relationships that may indicate potential threats. Figure \ref{fig:overview} shows how predictive algorithms are applied in the Cybercrime domain. In addition, this section also focuses on different datasets adopted by researchers to develop new models and discuss how different algorithms are used to predict cybercrime.

\subsection{Cybercrime Datasets} \label{sec:Dataset}

As part of this research, we collected publicly available datasets that were used by researchers in this domain and listed them in table \ref{tab dataset}. These datasets cover a range of data types, including bytecode images, network traffic, emails, and URLs. Researchers use these datasets to develop and evaluate ML models for cybercrime prediction, such as identifying malware, detecting fraudulent emails, and predicting phishing attacks. Some of the datasets listed in the table \ref{tab dataset} include the malware classification dataset, which contains bytecode images of various types of malware; the Virtual Private Network (VPN) dataset and The Onion Router (TOR) dataset, which contain network traffic data related to VPN and TOR connections, respectively; and the Enron dataset, which contains email data from the Enron Corporation. 

These datasets have been used to develop and evaluate ML models for malware detection, VPN and TOR detection, and fraudulent email detection. Other datasets listed in the table \ref{tab dataset} include the intrusion detection system(IDS) dataset, which contains network traffic data for intrusion detection; the ransomware dataset, which contains text data related to ransomware attacks; and the phishing website dataset, which contains Uniform Resource Locators (URLs) related to phishing attacks. 

\begin{longtable}{L{0.8\textwidth}L{0.2\textwidth}}
    \caption{Datasets used in Cybercrime Prediction}\label{tab dataset}
    \\ \hline
    Dataset Link                                                                                  & Dataset Type           \\ \hline
    https://github.com/AFAgarap/malware-classification/tree/master/dataset \cite{rustam2023malware,kumar2021mcft,kumar2022dtmic,go2020visualization}                     & Bytecode - Images       \\ \hline
    https://www.unb.ca/cic/datasets/vpn.html \cite{singh2021deep}                                                     & Network Traffic - Text \\ \hline
    https://www.unb.ca/cic/datasets/tor.html \cite{singh2021deep}                                                     & Network Traffic - Text \\ \hline
    https://monkey.org/$\sim$jose/wiki/doku.php \cite{gogoi2022phishing}                                               & Email - Text           \\ \hline
    https://www.kaggle.com/datasets/rtatman/fraudulent-email-corpus \cite{gogoi2022phishing}                              & Email - Text           \\ \hline
    https://www.cs.cmu.edu/$\sim$enron/ \cite{gogoi2022phishing}                                                           & Email - Text           \\ \hline
    https://www.kaggle.com/competitions/malware-classification/data \cite{kumar2021mcft,kumar2022dtmic}                               & Bytecode - Images      \\ \hline
    https://web.cs.hacettepe.edu.tr/$\sim$selman/malevis/ \cite{alodat2021detection}                                         & Bytecode - Images      \\ \hline
    https://www.unb.ca/cic/datasets/nsl.html \cite{zhao2019transfer,klein2022jasmine,klein2021plusmine,ravi2022recurrent,vinayakumar2019deep}                                                   & Network Traffic - Text \\ \hline
    https://www.honeynet.org/category/honeypot/ \cite{kumar2022dtmic}                                                & Bytecode - Images      \\ \hline
    https://www.ll.mit.edu/r-d/datasets/2000-darpa-intrusion-detection-scenario-specific-datasets \cite{chadza2020learning} & Network Traffic - Text \\ \hline
    https://www.unb.ca/cic/datasets/ids-2018.html \cite{chadza2020learning}                                                 & Network Traffic - Text \\ \hline
    https://research.unsw.edu.au/projects/unsw-nb15-dataset \cite{klein2022jasmine,klein2021plusmine,zaman2022trustworthy,ravi2022recurrent,vinayakumar2019deep}                                       & Network Traffic - Text \\ \hline
    https://github.com/PSJoshi/Notes/wiki/Datasets \cite{khan2020digital}                                                & Ransomware - Text      \\ \hline
    https://github.com/ebubekirbbr/pdd/tree/master/input \cite{chatterjee2019detecting}                                         & URL                    \\ \hline
    https://www.phishlabs.com/covid-19-threat-intelligence/ \cite{mvula2022covid}                                      & URL                    \\ \hline
    https://www.unb.ca/cic/datasets/index.html  \cite{ahammad2022phishing}                                                  & URL                    \\ \hline
    Scammer.info \cite{chen2021ai}                                                                                 & Web Crawl - Text      \\ \hline
    Urlscan.io \cite{chen2021ai}                                                                                    & Web Crawl - Text       \\ \hline
    https://www.kaggle.com/datasets/akashkr/phishing-website-dataset  \cite{mridha2021phishing}                            & URL                    \\ \hline
    https://sel.psu.edu.sa/Research/datasets/2016\_WSN-DS.php. \cite{ravi2022recurrent,vinayakumar2019deep}                                    & IDS - Numeric          \\ \hline
    http://www.takakura.com/Kyoto\_data/ \cite{vinayakumar2019deep}                                                         & IDS - Numeric          \\ \hline
    https://nlp.amrita.edu/DMD2018/ \cite{akarsh2019deep}                                                             & URL                    \\ \hline
    https://www.kaggle.com/datasets/drkhurramshahzad/ violent-views-detection-dataset-in-urdu \cite{akram2021violent}      & Text                   \\ \hline
    https://web.archive.org/web/20081219063350/ http://www.cernet2.edu.cn/index\_en.htm \cite{sun2020deepdom}            & URL                    \\ \hline
    http://5000best.com/websites \cite{wazirali2021sustaining}                                                                & Legitimate -URL        \\ \hline
    https://www.phishtank.com/ \cite{wazirali2021sustaining}                                                                  & Phishing - URL         \\ \hline
    https://www.unb.ca/cic/datasets/ids-2017.html \cite{ravi2022recurrent,vinayakumar2019deep}                                                & Network Traffic -      \\ \hline
\end{longtable} 

\subsection{Machine Learning for Cybercrime Prediction} \label{subsec:ML}

In recent years, rapid technological advancements have prompted a shift from traditional to electronic methods of living. This transition has attracted cybercriminals who exploit the Internet for activities like phishing, aimed at acquiring sensitive personal information. The COVID-19 pandemic exacerbated this situation, leading to a surge in pandemic-related cyberattacks. As cyber threats persist, continuous innovation is imperative for cybersecurity experts. Despite existing anti-phishing measures, such as blocklists and heuristics, proving insufficient, research has been directed towards predictive solutions. Section 4.2 focuses on ML techniques for cybercrime prediction, with Table \ref{tab ML} listing key contributions in this domain.

\small
\begin{longtable}{L{0.25\textwidth}L{0.35\textwidth}
   L{0.2\textwidth}}
    \caption{Machine Learning approaches in Cybercrime Prediction}\label{tab ML}
    \\ \hline
    Existing Research   & Algorithms & Performance      \\ \hline
    
       Kumari et al.  \cite{kumari2018machine}                                                                                                                                                                   & Naive Bayes Classification with NLTK                                                                & Accuracy - 77\%                            \\ \hline
 Mvula et al. \cite{mvula2022covid}         & DT, RF, GBM, XGBoost, and SVM                                                     & Accuracy - 97.93\%                         \\ \hline
  Shah et al.  \cite{shah2019compromised}                                                                                                                                                                   & Knowledge-based algorithm that focuses on clustering method for pattern detection and identification & Accuracy - 99\% \& Sensitivity - 90\%      \\ \hline
    Balakrishnan et al. \cite{balakrishnan2020improving}                                                                                     & Naive Bayes (NB) and RF  & Accuracy - 92\%                            \\ \hline
   Ahammad et al \cite{ahammad2022phishing}                                                                            & LightGBM, RF, DT, LR and SVM                                                                         & Accuracy - 86\%                            \\ \hline
  Chen et. al \cite{chen2021ai} &  LightGBM  & Accuracy - 98\% and F1 Score - 97.95\%     \\ \hline
   Wang et. al \cite{wang2022ssappidentify}                                                                            & RF, SVM, GNB (Gaussian Naive Bayes)                                                                  & Accuracy - 94.5\%                          \\ \hline
  Oh et. al  \cite{oh2020wipimization} & DT, SVM, RF                                                                                         & Accuracy - 95\%                            \\ \hline

   Mridha et. al \cite{mridha2021phishing}                                                                                                                               & ANN and RF                                                                                           & Accuracy - 99\%                            \\ \hline
   Palad et. al \cite{palad2019document}  & DT, NB, and Sequential Minimal Optimization & Accuracy - 79\%                          \\ \hline
\end{longtable}
Mvula et al. \cite{mvula2022covid}, minimal-feature ML techniques distinguish valid from malicious COVID-19-related domains, highlighting lexical functions and subdomains. Addressing security in digital transformation, Shah et al.\cite{shah2019compromised} introduces enterprise-level CUC detection via user behavior and a KBS, achieving 99\% accuracy. Balakrishnan et al.\cite{balakrishnan2020improving} explores psychological features' link to cyberbullying, presenting an automatic detection tool using Twitter users' attributes. Sentiments and personalities, but not emotions, aid cyberbullying detection. Extraversion, neuroticism, agreeableness, and psychopathy are influential traits.

Ahammad et al. \cite{ahammad2022phishing}, ML algorithms detect malicious URLs based on their characteristics and behaviors. To combat evasion, ML models with NLP, DT, LR, SVM, and Light GBM classify URLs. 
Mridha et al.\cite{mridha2021phishing} employs RF and ANN-based ML, achieving up to 99\% accuracy in identifying phishing URLs.

Another type of cybercrime is, Technical Support Scam (TSS), which affects the homeowner's property. Chen et al. introduces \cite{chen2021ai}, an AI@TSS system, based on LightGBM, is developed to detect Technical Support Scam (TSS) cybercrimes. They gathered 8263 TSS web page samples and 8\,263 malicious web page samples, using 42 functions for modeling. Results show AI@TSS achieves 98\% accuracy and 100\% precision, outperforming existing methods. Oh et al. \cite{oh2020wipimization} address the misuse of data wiping, proposing an anti-anti-forensic method using NTFS transaction features and machine learning. This approach identifies wiped files efficiently and provides insights into the data-wiping process. In \cite{palad2019document} study, the Weka text mining tool is used to classify an online scam dataset with 14,098 Filipino words. J48 Decision Tree outperforms other classifiers, achieving the highest accuracy and lowest error, validated through responses.

In another study \cite{wang2022ssappidentify}, the focus is on Shadowsocks, a commonly used method for bypassing firewalls. The authors developed a traffic identification system for applications over Shadowsocks, enhancing monitoring and evidence gathering for cybercrime activities. They incorporated a sliding window JS divergence feature into the system, maintaining application features while reducing the impact of smartphone variations without compromising accuracy. In a separate investigation \cite{kumari2018machine}, the authors utilized two training datasets: one sourced from Facebook and Twitter using the Facepager software tool and another from online sources. Their objective was to extract cybercrime data, create labeled classes (positive and negative), and preprocess it for supervised machine learning. Employing Natural Language Toolkit (NLTK) and Scikit-learn, the authors achieved high classifier accuracy and text classification confidence values for cybercrime data, demonstrating improved dataset accuracy.

Furthermore, A multimodal approach in cybercrime detection combines diverse data types such as text, images, network traffic, and behavior patterns. It improves accuracy by cross-verifying suspicious activities, aids in contextualizing threats, and detects complex attacks. This approach analyzes user and system behaviors, attributes threats effectively, and acts as an early warning system. It also offers robust defense mechanisms, adapts to evolving threats, and reduces the attack surface. For instance, \cite{gautam2023automatic, gautam2023email, gautam2022effect, gautam2022performance} have employed a multimodal mechanism within the machine learning framework to enhance automated cybercrime detection. They further stated that an intelligent cybercrime detection system is needed to automatically address and identify disturbing cybercrime incidents on social media platforms.

ML has great potential in cybercrime detection, but challenges persist. Traditional methods often fail to identify complex attacks like zero-days and advanced persistent threats (APT). Incorporating advanced DL techniques, can enhance content analysis and prediction accuracy. The imbalance in cybercrime datasets can lead to overfitting, while concerns about model transparency and interpretability are growing. In this context, ongoing research is essential to evaluate various ML algorithms, explore effective feature engineering techniques, and investigate the potential of hybrid models that combine multiple approaches. Addressing these issues will significantly advance cybercrime prediction and prevention.

\subsection{Deep Learning for Cybercrime Prediction} \label{subsec:DL}

An increasingly prevalent method for detecting cybercrime is DL. Due to the exponential rise of cyber threats and attacks, more than traditional prediction techniques are required to quickly identify and mitigate new and unidentified threats. DL algorithms can scan vast amounts of data, find obscure patterns, and automatically pick up on and respond to new threats. Moreover, APTs and zero-day attacks, which are frequently missed by conventional prediction techniques, may be identified and recognized using DL. Deep learning's use in cybercrime prediction has sparked the creation of sophisticated prediction models for a range of online threats, such as malware, phishing, botnets, and domain-
generation algorithms (DGA)-based attacks. In table \ref{tab DL}, we list the important research performed in this area and discuss more in detail below.
 \small
\begin{longtable}{L{0.3\textwidth}L{0.35\textwidth}
   L{0.2\textwidth}}
    \caption{Deep Learning approaches in Cybercrime Prediction}\label{tab DL}
    \\ \hline
    Existing Research                                                                                                & Algorithms                                             & Performance                                       \\ \hline
  Sun et al. \cite{sun2020deepdom}                                                                                                                                                                                                                                                                      & DeepWalk, Metapath2Vec, GraphSAGE and SHetGCN             & Accuracy: - 97\%                                  \\ \hline
   Ngejane et al. \cite{ngejane2021digital}                                                                                                                                                                                                                                                                & LR, XGBoost, MLP \& BiLSTM                                & Accuracy:- 98 F1 Score :-  70\%                   \\ \hline
  Wazirali et al. \cite{wazirali2021sustaining}                                                                                                                                                                                                                                                & Feature Selection CNN (FS-CNN)                            & Accuracy:- 99\%                                   \\ \hline
   Ravi et al. \cite{ravi2022attention}                                                                                                                                                                                                 & CNN, LSTM, BLSTM                                          & Accuracy: - 95\%    
\\ \hline
  Adebowal et al. \cite{adebowale2019deep}                                                                                                                                                                                                             & CNN+LSTM                                                  & Accuracy: - 93.28 
\\ \hline
  Ravi et al.   \cite{ravi2022recurrent}                    & NB, LR, KNN, DT, RF, RNN, LSTM, GRU and Proposed Approach & Accuracy:- \textgreater 95\% for all 5 data sets                                 \\ \hline
 Yuan et al.  \cite{yuan2019stealthy}                                                                                                                                                                                         & R-CNN                                                                                                   & Accuracy: - 85.00\%   
    \\ \hline
  Ravi et al.  \cite{ravi2021adversarial} & CNN, LSTM, GRU, BLSTM, RNN & Accuracy: - 99\%                                                               \\ \hline
 Vinayakumar et al.  \cite{vinayakumar2019deep}                                                                                                                                                                                                                                                                     & LR, NB, KNN, DT, AB, RF, SVM and DNN                            & Accuracy: - 93.50\%                               \\ \hline
   Vinayakumar et a. \cite{vinayakumar2019robust}                                                                                                                                                                                                                            & CNN, LSTM                                                 & Accuracy: - 96.30\%                                                         \\ \hline
 Akarsh et al.  \cite{akarsh2019deep}                                                                                                                                                                                             & DGA, LSTM                                                                                                & Accuracy: - 98.7\%                                \\ \hline
   Akram et al.  \cite{akram2021violent}                                                                                                                                                                                                                        & CNN, LSTM                                                                                       & F1 score :- 88.1\%                                \\ \hline
\end{longtable}

To identify malicious domains in the DNS environment, Sun et al. \cite{sun2020deepdom} have created an intelligent system called DeepDom. To capture various entities, they used a Heterogeneous Information Network (HIN), and to categorize domain nodes, they suggested a brand-new GCN approach called scalable and heterogeneous Graph Convolutional Network (SHetGCN). SHetGCN handles node characteristics and structural information and supports inductive node embedding using meta-path-based short random walks to direct convolution processes. Further, Ravi et al. \cite{ravi2021adversarial} have created a DeepDom prototype and verified its efficacy through extensive tests utilizing DNS information gathered from CERNET2 a second generation of China Education and Research Network. For traditional reverse engineering approaches, employing domain-generation algorithms (DGAs) by cybercriminals to escape blocklisting or server shutdown poses a significant difficulty. These methods take time, are prone to mistakes, and have restrictions. Hence, a real-time, automated method with a high prediction rate is required. A unique method for detecting DNS homograph attacks and randomly created domain names using DL without reverse engineering or NXDomain inspection is presented by Ravi et al. \cite{ravi2021adversarial}. The study underlines the need for more reliable prediction models to combat adversarial learning.

Wazirali et al. \cite{wazirali2021sustaining}, the authors propose a solution to enhance phishing website identification, addressing issues like low accuracy, steep learning curves, and compatibility with low-power embedded technology. Their approach utilizes a CNN algorithm, Clustering and Feature Methods, and Software Defined Network (SDN) technologies, classifying URLs based on sequential patterns and URL metadata without accessing website content or third-party services. Similarly, Adebowale et al. \cite{adebowale2019deep}, authors develop a DL-based phishing detection system incorporating website content, graphics, frame elements, and URLs. They use CNN and LSTM algorithms to classify phishing websites. Ravi et al. \cite{ravi2022recurrent} introduces an end-to-end network attack detection and classification model, employing DL-based recurrent models with kernel-based principal component analysis (KPCA) for feature selection and an ensemble meta-classifier for data classification. Additionally, Vinayakumar et al. \cite{vinayakumar2019deep} emphasize the importance of regularly updated malware datasets for creating a flexible intrusion detection system (IDS). They propose a deep neural network (DNN) based IDS, adaptable to changing network behavior and emerging attack techniques. The authors evaluate various datasets to determine optimal algorithms and introduce a highly scalable hybrid DNN architecture, "scale-hybrid-IDS-AlertNet," for proactive threat detection through continuous monitoring of network traffic and host-level events.

The escalating malware threat has spurred extensive research. Conventional real-time detection methods like static and dynamic analysis suffer from sluggishness. Deep Learning (DL) offers an efficient alternative, eliminating labor-intensive feature engineering. Vinayakumar et al. \cite{vinayakumar2019robust}, DL takes the spotlight for zero-day malware detection via innovative image processing. This paves the way for scalable hybrid DL methods for reliable detection. Ravi et al. \cite{ravi2022attention} addresses malware in Internet of Medical Things (IoMT) devices, advocating automated detection. They propose cross-architecture attention-based DL for IoMT malware detection, automating CPU architecture detection and feature extraction from ELF files. DL's rapid advancements expose vulnerabilities, notably in adversarial learning. Adversaries exploit these vulnerabilities to evade image-based detection in the real world. Yuan et al. \cite{yuan2019stealthy} explores adversarial promotional porn pictures (APPIs) used for illicit advertising. They introduce Malena, a DL-based approach revealing obscured sexual content areas in these images. Malena sheds light on real-world adversarial images and their clandestine use.

While DL algorithms have shown great promise in handling some of the challenges traditional ML methods face in cybercrime detection, some drawbacks still need to be addressed. One of the significant challenges is the need for large amounts of labeled data to train the models effectively. As cybercrime datasets are often imbalanced, with most cases being benign or non-malicious, this can lead to biased models and inaccurate predictions. Another challenge is the computational cost of training DL models, which requires significant computational resources and can be time-consuming. To address these challenges, TL can be used to improve the performance and efficiency of DL models for cybercrime detection. 

\subsection{Transfer Learning for Cybercrime Prediction} \label{subsec:TL}

Transfer learning is a powerful technique in ML that involves leveraging knowledge and pre-trained models to improve the performance of a new model on a different but related task. It has been used successfully in many fields, including computer vision, NLP, and speech recognition. Recently, TL has also shown promise in the field of cybersecurity, particularly in the area of cybercrime prediction. Cybercrime prediction aims to detect and prevent malicious activities on computer networks before they occur. Traditional methods of cybercrime prediction rely on signature-based approaches, where known threats are identified by matching their characteristics to a database of known malware signatures. However, these methods could be improved in their ability to detect new and evolving threats, which require significant effort and resources to identify and analyze.

\begin{figure*}[htbp!]
    \begin{center}
        \includegraphics[scale=0.5]{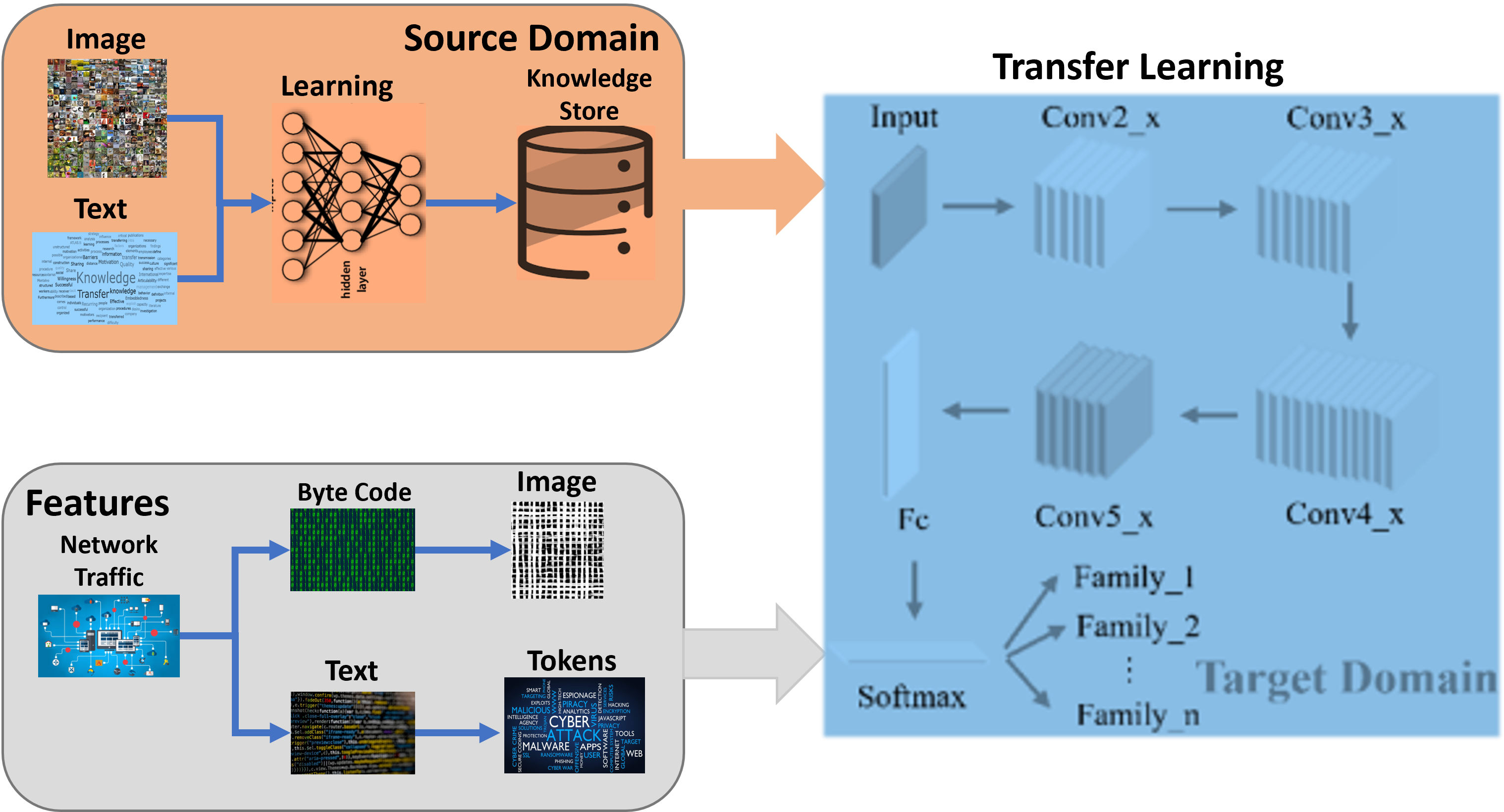}
        \caption{\centering{A pictorial view of transfer learning architecture in cybercrime predictions}}
        \label{fig:TL}
    \end{center}
\end{figure*}

Transfer learning, as illustrated in Figure \ref{fig:TL}, offers a promising solution to enhance cybercrime prediction accuracy. It involves repurposing pre-trained models, like using a network traffic model for anomaly detection in cyber attacks. Additionally, transfer learning can tap into models trained in fields like NLP or computer vision to uncover patterns in cybercrime activities, such as text analysis for malicious indicators in emails or chats. This approach not only boosts prediction accuracy but also mitigates the challenge of limited labeled data in cybersecurity. Our focus in this work is on 13 recent TL-related research articles, detailed in Table \ref{tab TL} below.

Rustam et al. \cite{rustam2023malware} related to the prediction of malware using TL, researchers developed a bimodal approach where the features are extracted using VVG and ResNet models and were fed into ML models for predicting classifier probability. These classification probabilities are then fed into the final Ml model that makes predictions. Singh et al. \cite{singh2021deep} focusing on Darknet, researchers proposed a deep TL based method to transform network traffic numerical data into image data, and TL based features were then fed into a bi-level classifier to classify malicious activity. In this study, the author achieved an accuracy of 96$\%$. Similar to this study, another study \cite{kumar2021mcft} transformed bytecode data into images and fed them to an MCFT-CNN that gets low-level features from ImageNet to classify malware. A similar study \cite{alodat2021detection} that focuses on pre-trained CNN algorithms inputs image datasets to classify malware. DTMIC \cite{kumar2022dtmic} is also a method based on earlier concepts where binary data is converted to images and fed into ImageNet for classifying malware. Trustsign \cite{nahmias2019trustsign} is another malware detection method that uses image data to be fed into VGG19, in which a max-pooling layer for malware classification replaced three fully connected layers. Apart from these, there are multiple other studies \cite{go2020visualization,phoka2019image,yadav2022efficientnet} that focuses on TL for malware and phishing detection by inputting original image or image generated from bytecodes.
\small
\begin{longtable}
{L{0.2\textwidth}L{0.4\textwidth}
   L{0.2\textwidth}}
    \caption{Transfer Learning approaches in Cybercrime Prediction}\label{tab TL}
    \\ \hline
  Existing Research                                                                                                                                                                                                                                                                               & Algorithms                                                                                                                                               & Performance                             \\ \hline
   Rustam et al. \cite{rustam2023malware}                                             & VVG-16, ResNet-50, SVC, RF, KNN, LR, CNN, InceptionV3 and EfficientNetB0                                                                                           & Aggregate Accuracy - 100\% 
        \\ \hline
   Kumar et al.  \cite{kumar2021mcft}                                         & Malware Classification Fine Tune-CNN \& ImageNet                                                                                                                   & Accuracy - 99.18\%                  \\ \hline
  Nahmias et al.  \cite{nahmias2019trustsign}                                                                                                          & Modified VGGNet                                                                                                                                                    & Accuracy - 99.5\%   
    \\ \hline
   Hou et al. \cite{hou2022identification}                                                                                                      & F1 Score - 89.66\%  
    \\ \hline
   Singh et al. \cite{singh2021deep}                                                                           & ResNet(18, 50, 101), VGG(16, 19), AlexNet, DenseNet, GoogleNet, InceptionV3, SqueezeNet, SVC, DT and RF                                                            & Accuracy - 96\%                      \\ \hline
  Gogoi et al. \cite{gogoi2022phishing}                                                                                                               & BERT and DistilBERT                                                                                                                                                & F1 Score - 99\%                    \\ \hline
   Alodat et al. \cite{alodat2021detection}                                                                                         & MobileNetV2, InceptionV3, ResNet50, LittleVGG                                                                                                                      & Accuracy - 95\%                      \\ \hline
   Zhao et al. \cite{zhao2019transfer}                                                                             &  Cluster Enhanced Transfer Learning (C2HTL), SVC and Heterogeneous Map (HeMAP)                                             & Best Accuracy - 88\%                 \\ \hline
  Kumar et al.  \cite{kumar2022dtmic}                                                                             & VGG(16, 19), ResNet-50, InceptionV3 and DTMIC                                                                                                                      & Aggregate Accuracy - 95\%                       
     \\ \hline
   Go et al. \cite{go2020visualization}                                                                                                                                                                                              & ResNeXt                                                                                                                                                             & Accuracy - 98.32\% and 98.86\%   
           \\ \hline
Yadav et al.        \cite{yadav2022efficientnet}                                                                                                                                                                                       & Multiple CNN pre-trained models                                                                                                                                   & 95\% Accuracy EfficientNet-B4      
    \\ \hline
  Phoka et al.  \cite{phoka2019image}         & Inception (V3, V4), ResNet (V1, V2 and Inception)                                                                                                                  & Best Accuracy - 97\%  
    \\ \hline
  Chadza et al.  \cite{chadza2020learning}                    & Conventional Machine Learning (CML) HMM, Baum Welch (BW), Viterbi training (VT), differential evolution (DE), gradient descent (GD), and simulated annealing (SA) & Best Accuracy - 96.3\% 
                      \\ \hline

\end{longtable}

Apart from image based transfer learning, researchers also focused on cluster and Markov model based cybercrime detection. Zhao et al. \cite{zhao2019transfer} focuses on unknown network attacks uses k-mean clustering based transfer learning to generate target domain based clusters. The Euclidean based similarities were computed between source and target domains to be fed into the classifier to identify unknown network attacks. Another study \cite{chadza2020learning} that focuses on Hidden Markov Models uses alerts generated by Snort Intrusion Detection System (IDS) to detect the current and future state of network attacks. This model performed with a high accuracy of 96.3$\%$. Apart from image and data based TL, We also looked at NLP-based methods in TL. Gogoi et al. \cite{gogoi2022phishing} focused on phishing email detection and used BERT and DistilBERT to classify phishing and normal email by feeding these algorithms with tokens generated from subject and email text. Another study \cite{hou2022identification} focusing on detecting Chinese jargon used TL on telegram chat data. This data is fed into three feature extraction methods based on lexical, vector based on TL, and dictionary methods. An outlier detection method is employed on these features to identify the jargon.

While Transfer Learning (TL) offers significant benefits for enhancing Deep Learning (DL) models, it comes with certain limitations that can be mitigated through Active Learning (AL) and Reinforcement Learning (RL). A key challenge is the requirement for relevant pre-trained models, particularly when dealing with novel data. AL addresses this by selecting informative samples for labeling, reducing the need for extensive labeled data. TL also demands time-consuming fine-tuning, which RL can optimize through a reward-based approach, reducing the computational burden. Moreover, the risk of negative transfer, where pre-trained knowledge isn't pertinent, can be managed using AL and RL techniques, as discussed in subsection \ref{subsec:ESL}.

\subsection{Early Stage Algorithms in Cybercrime Prediction} \label{subsec:ESL}

\begin{figure*}[htbp!]
    \begin{center}
        \includegraphics[scale=0.4]{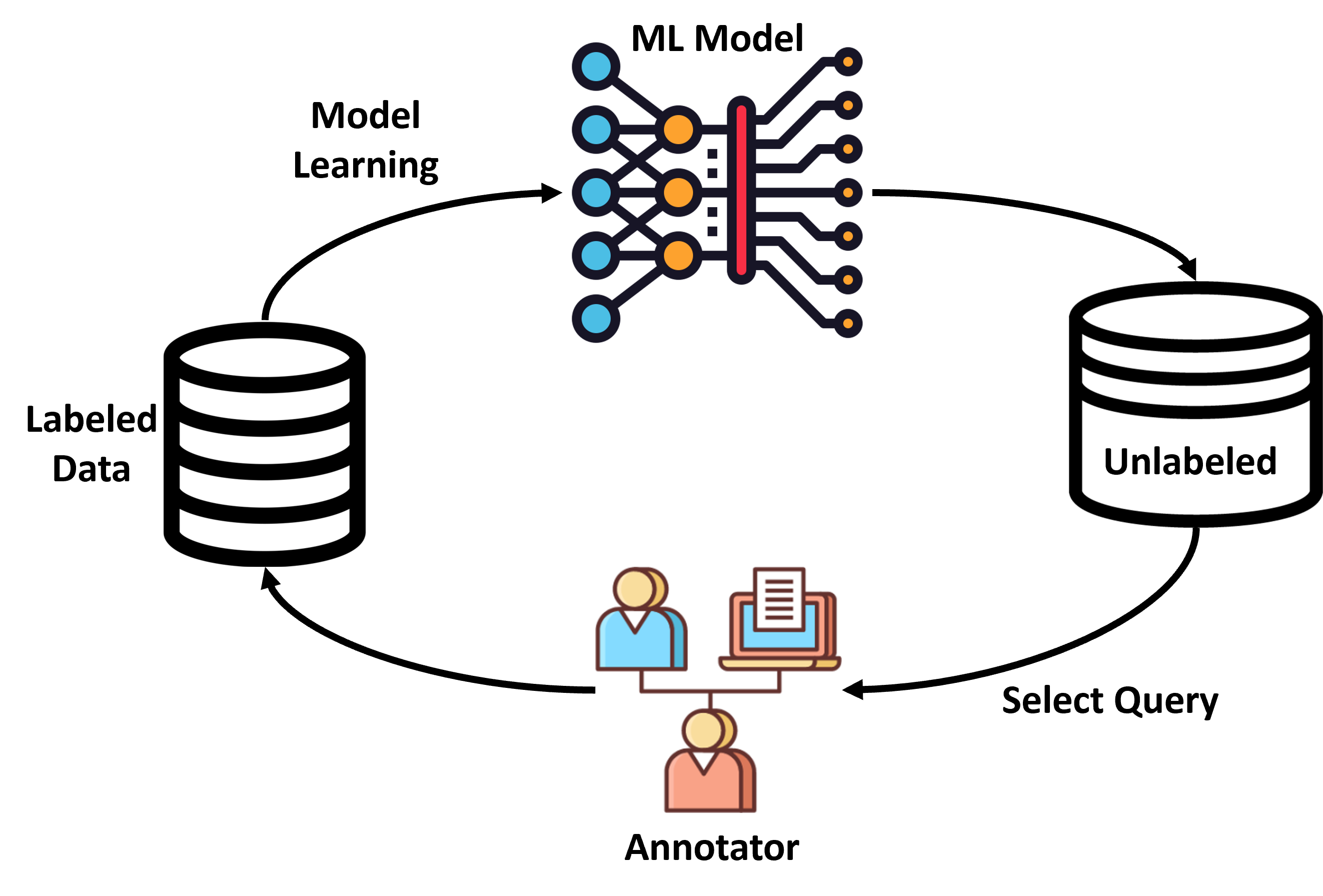}
        \caption{\centering{A general pictorial view of Active Learning approach}}
        \label{fig:AL}
    \end{center}
\end{figure*}

In this article, AL and RL are categorized as part of early-stage algorithms, as research on AL and RL in cybercrime prediction is still in the early stages. AL is an ML technique that can significantly improve the efficiency and effectiveness of predictive models by selecting the most informative data samples for annotation. In AL, the algorithm iteratively selects a subset of unlabeled data points most likely to provide valuable information to a model's decision-making process. These data points are then labeled by a human expert and added to the training set, allowing the model to improve its accuracy with each iteration, as shown in figure \ref{fig:AL}. This approach can be beneficial in domains where labeled data is insufficient or costly to obtain, such as in cybercrime prediction. 

Predicting cybercrime incidents accurately is a formidable task due to the ever-evolving nature of cyber threats and the scarcity of labeled data for training predictive models. Active Learning (AL) emerges as a valuable tool in overcoming these challenges, as it reduces the need for labeled data and enhances the performance of predictive models. One notable application of AL in cybercrime prediction is the Jasmine system \cite{klein2022jasmine}, which leverages uncertainty and anomaly scores to assess the suitability of observations for querying. It features dynamic updating, allowing the model to optimize its querying strategy based on uncertainties and anomalies, consistently delivering robust results. A recent advancement beyond Jasmine is the introduction of Plusmine, a network intrusion detection system that combines AL and semi-supervised learning. Plusmine builds upon Jasmine, showcasing superior performance across various dataset configurations.

Another significant development in this realm is the DNAact-Ran method \cite{khan2020digital}, which proposes digital DNA sequencing for ransomware detection. This approach employs a combination of AL and semi-supervised learning techniques to enhance model training. AL selects samples from unlabeled data for manual labeling, while semi-supervised learning harnesses the available limited labeled data and abundant unlabeled data to further improve the model's performance. These innovative approaches highlight AL's potential to strengthen cybercrime prediction and network intrusion detection.

\begin{figure*}[htbp!]
    \begin{center}
        \includegraphics[scale=0.45]{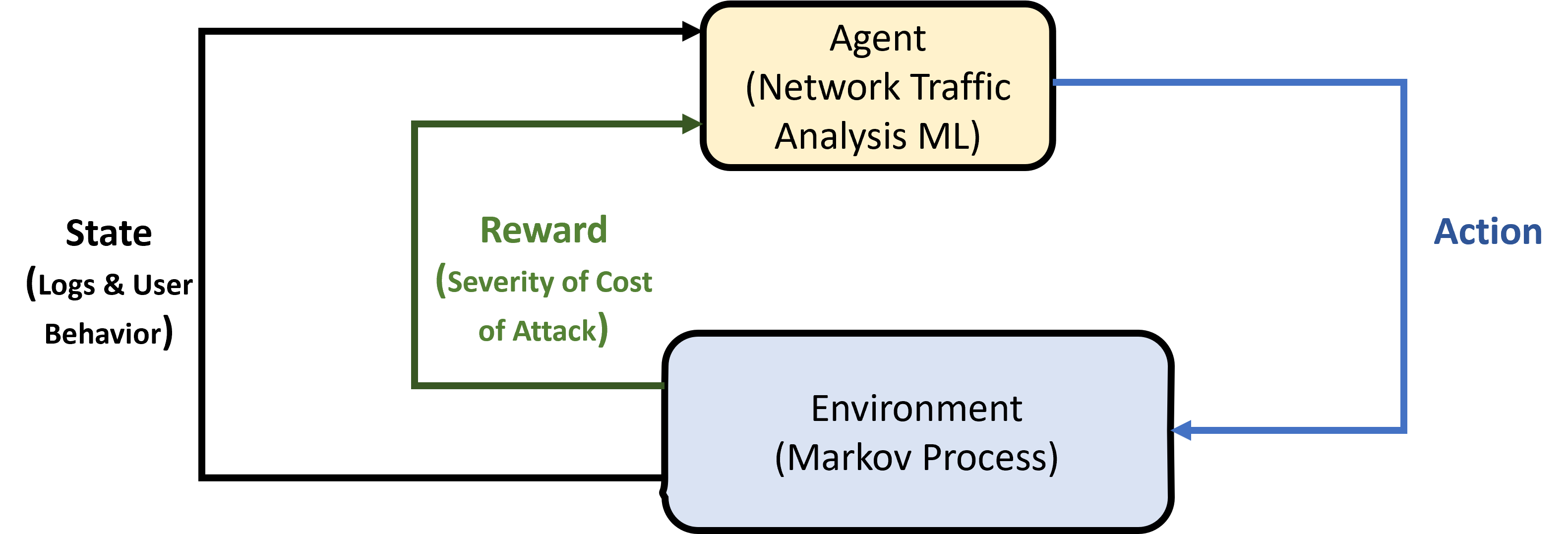}
        \caption{\centering{An architectural view of reinforcement learning in cybercrime prediction}}
        \label{fig:RL}
    \end{center}
\end{figure*}

\textbf{Reinforcement learning} is an ML technique gaining popularity in cybercrime prediction. RL is a type of artificial intelligence where the machine learns to make decisions by interacting with its environment, and it is commonly used in gaming and robotics. The same concept can be applied to cybercrime prediction, where the machine learns to predict and respond to cyber threats based on its interaction with the environment, as shown in figure \ref{fig:RL}.

In cybercrime prediction, RL can be used to train machines to learn from past cyber attacks and adapt their behavior accordingly to detect and respond to similar attacks in the future. RL can help identify patterns of cybercrime behavior, which can be used to develop better predictive models. One potential application of RL in cybercrime prediction is identifying malicious network traffic. Chatterjee et al. \cite{chatterjee2019detecting} proposed a RL framework for automated URL-based phishing detection. RL's deep neural network-based implementation to map the sequential decision-making process complements existing phishing detection methodologies and offers a more dynamic and self-adaptive phishing identification system. This work is a foundation for a more efficient framework but is not yet optimized for real-world implementation. A recent study that focused on RL \cite{zaman2022trustworthy} proposed a security mechanism to detect cyberattacks in IoT devices by employing RL. The proposed method uses the DeepQ algorithm to train the agent, which takes the lexical features of URLs as input, can dynamically adapt to new phishing attacks, and achieves an average accuracy of 97.29$\%$ using the UNSW-NB dataset. The results demonstrate that the proposed study has the potential to be deployed as a security mechanism against cybercrimes. However, RL in cybercrime prediction is still in its early stages, and some challenges must be addressed.

\section{Future Research Directions} 
\label{sec:FRD}

Based on the identified research gaps and insights from reviewed articles, we suggest future research directions in the domain of cybercrime prediction.

\textbf{Developing hybrid models and Incorporating context-awareness}: Future research can explore the development of hybrid models that combine the strengths of different machine, deep, and TL techniques. These models could offer better performance, Interpretability, and adaptability than individual techniques. Cybercriminal activities often have contextual information that can provide valuable insights for prediction. Future research should incorporate context awareness \cite{al2021lightweight} in prediction models to improve their effectiveness in detecting and preventing cybercrimes.

\textbf{Leveraging emerging technologies}: Researchers can explore the integration of emerging technologies, such as edge computing \cite{pan2018cybersecurity}, federated learning, and blockchain \cite{maleh2020blockchain}, to enhance the performance, privacy, and security of cybercrime prediction models. For instance \cite{alazab2021federated}, federated learning can enable collaborative learning across multiple organizations while preserving data privacy.

\textbf{Cross-disciplinary Approaches and Ethical Considerations}: Cybercrime prediction can benefit from cross-disciplinary approaches, incorporating insights from fields such as psychology, criminology, and social network analysis \cite{holt2021researching,custers2021profiling}. Researchers can develop more comprehensive and effective models for predicting and mitigating cybercriminal activities by integrating knowledge from these domains. As cybercrime prediction models become more advanced and pervasive, it is also essential to consider their ethical implications \cite{ang2021legal,hughes2021too}. Future research should address issues such as potential bias in training data, privacy concerns, and the impact of false positives and negatives on individuals and organizations.

\textbf{Active learning $\&$ Reinforcement learning for cybercrime prediction}: Delving deeper into AL techniques for cybercrime prediction models holds promise for several reasons. Since AL focuses on obtaining labels for the most informative data points, it allows for more efficient use of expert resources. This efficiency can be crucial in cybercrime prediction, where domain experts are often in high demand and short supply. By minimizing expert intervention, AL can accelerate the development of accurate prediction models. Cybercrime prediction often involves class-imbalanced datasets, where the number of malicious instances is significantly lower than that of benign ones. AL techniques can help address this issue by prioritizing the acquisition of labels for underrepresented classes, thereby improving the model's ability to detect rare or emerging threats. Finally, AL can help prediction models adapt to these changes by selectively incorporating new, informative data into the training process. This continuous learning approach allows models to remain relevant and effective even as cyber threats evolve.

By exploring these future research directions, researchers can contribute to developing more accurate, efficient, and adaptive models for predicting and preventing cybercriminal activities. These advancements will not only enhance the effectiveness of existing cybercrime prediction models but also pave the way for developing novel techniques that address the challenges posed by cyber threats.

\section{Conclusion} 
\label{sec:CC_Conc}

In conclusion, this survey paper has thoroughly examined the recent advancements in cybercrime prediction using machine, deep, and transfer learning techniques. We have highlighted the critical innovations that have propelled the field forward, such as applying deep learning models for feature extraction, transfer learning for leveraging pre-existing knowledge, and integrating active and reinforcement learning approaches for adaptive cyber defense. To support future researchers, we gathered and provided publicly available datasets used for cybercrime prediction in this work. We have also discussed the research gaps, including the need for causal reasoning, cross-disciplinary approaches, and ethical considerations in developing these models. In addition, we have outlined several future research directions, such as exploring active learning and reinforcement learning techniques, incorporating causal reasoning, and addressing the ethical implications of cybercrime prediction models. By addressing these research gaps and pursuing innovative solutions, researchers can contribute to developing more accurate, efficient, and reliable models for predicting and preventing cybercriminal activities. Furthermore, these developments will enable the creation of adaptive and proactive cyber defense systems capable of responding to the ever-changing landscape of cyber threats. Ultimately, by focusing on these areas, the research community can help create a safer digital environment for individuals, organizations, and society.

\section*{Acknowledgement(s)}

The authors wish to acknowledge all those who contributed to the preparation and revision of the manuscript.

\section*{Disclosure statement}

The authors declare that there is no potential conflict of interest.

\section*{Funding}

The authors wish to acknowledge all those who contributed to the preparation and revision of the manuscript.

\bibliography{interactapasample}

\begin{thebibliography}{}

\bibitem[Adebowale et~al., 2019]{adebowale2019deep}
Adebowale, M.~A., Lwin, K.~T., and Hossain, M.~A. (2019).
\newblock Deep learning with convolutional neural network and long short-term memory for phishing detection.
\newblock pages 1--8.

\bibitem[Ahammad et~al., 2022]{ahammad2022phishing}
Ahammad, S.~H., Kale, S.~D., Upadhye, G.~D., Pande, S.~D., Babu, E.~V., Dhumane, A.~V., and Bahadur, M. D. K.~J. (2022).
\newblock Phishing url detection using machine learning methods.
\newblock {\em Advances in Engineering Software}, 173:103288.

\bibitem[Akarsh et~al., 2019]{akarsh2019deep}
Akarsh, S., Sriram, S., Poornachandran, P., Menon, V.~K., and Soman, K. (2019).
\newblock Deep learning framework for domain generation algorithms prediction using long short-term memory.
\newblock pages 666--671.

\bibitem[Akram and Shahzad, 2021]{akram2021violent}
Akram, M.~H. and Shahzad, K. (2021).
\newblock Violent views detection in urdu tweets.
\newblock pages 1--6.

\bibitem[Al-Hadhrami and Hussain, 2021]{al2021ddos}
Al-Hadhrami, Y. and Hussain, F.~K. (2021).
\newblock Ddos attacks in iot networks: a comprehensive systematic literature review.
\newblock {\em World Wide Web}, 24(3):971--1001.

\bibitem[Al-Mhiqani et~al., 2018]{al2018cyber}
Al-Mhiqani, M.~N., Ahmad, R., Yassin, W., Hassan, A., Abidin, Z.~Z., Ali, N.~S., and Abdulkareem, K.~H. (2018).
\newblock Cyber-security incidents: a review cases in cyber-physical systems.
\newblock {\em International Journal of Advanced Computer Science and Applications}, 9(1).

\bibitem[Al-Muhtadi et~al., 2021]{al2021lightweight}
Al-Muhtadi, J., Saleem, K., Al-Rabiaah, S., Imran, M., Gawanmeh, A., and Rodrigues, J.~J. (2021).
\newblock A lightweight cyber security framework with context-awareness for pervasive computing environments.
\newblock {\em Sustainable Cities and Society}, 66:102610.

\bibitem[Alazab et~al., 2021]{alazab2021federated}
Alazab, M., RM, S.~P., Parimala, M., Maddikunta, P. K.~R., Gadekallu, T.~R., and Pham, Q.-V. (2021).
\newblock Federated learning for cybersecurity: concepts, challenges, and future directions.
\newblock {\em IEEE Transactions on Industrial Informatics}, 18(5):3501--3509.

\bibitem[Alenezi et~al., 2020]{alenezi2020evolution}
Alenezi, M.~N., Alabdulrazzaq, H., Alshaher, A.~A., and Alkharang, M.~M. (2020).
\newblock Evolution of malware threats and techniques: A review.
\newblock {\em International journal of communication networks and information security}, 12(3):326--337.

\bibitem[Alhogail and Alsabih, 2021]{alhogail2021applying}
Alhogail, A. and Alsabih, A. (2021).
\newblock Applying machine learning and natural language processing to detect phishing email.
\newblock {\em Computers \& Security}, 110:102414.

\bibitem[Aljabri and Mirza, 2022]{aljabri2022phishing}
Aljabri, M. and Mirza, S. (2022).
\newblock Phishing attacks detection using machine learning and deep learning models.
\newblock pages 175--180.

\bibitem[Alkhalil et~al., 2021]{alkhalil2021phishing}
Alkhalil, Z., Hewage, C., Nawaf, L., and Khan, I. (2021).
\newblock Phishing attacks: A recent comprehensive study and a new anatomy.
\newblock {\em Frontiers in Computer Science}, 3:563060.

\bibitem[Alodat and Alodat, 2021]{alodat2021detection}
Alodat, I. and Alodat, M. (2021).
\newblock Detection of image malware steganography using deep transfer learning model.
\newblock pages 323--333.

\bibitem[Ang, 2021]{ang2021legal}
Ang, B. (2021).
\newblock Legal issues and ethical considerations in cyber forensic psychology.
\newblock pages 233--249.

\bibitem[Apruzzese et~al., 2018]{apruzzese2018effectiveness}
Apruzzese, G., Colajanni, M., Ferretti, L., Guido, A., and Marchetti, M. (2018).
\newblock On the effectiveness of machine and deep learning for cyber security.
\newblock pages 371--390.

\bibitem[Aslan et~al., 2023]{aslan2023comprehensive}
Aslan, {\"O}., Aktu{\u{g}}, S.~S., Ozkan-Okay, M., Yilmaz, A.~A., and Akin, E. (2023).
\newblock A comprehensive review of cyber security vulnerabilities, threats, attacks, and solutions.
\newblock {\em Electronics}, 12(6):1333.

\bibitem[Balakrishnan et~al., 2020]{balakrishnan2020improving}
Balakrishnan, V., Khan, S., and Arabnia, H.~R. (2020).
\newblock Improving cyberbullying detection using twitter users’ psychological features and machine learning.
\newblock {\em Computers \& Security}, 90:101710.

\bibitem[Broadhead, 2018]{broadhead2018contemporary}
Broadhead, S. (2018).
\newblock The contemporary cybercrime ecosystem: A multi-disciplinary overview of the state of affairs and developments.
\newblock {\em Computer Law \& Security Review}, 34(6):1180--1196.

\bibitem[Chadza et~al., 2020]{chadza2020learning}
Chadza, T., Kyriakopoulos, K.~G., and Lambotharan, S. (2020).
\newblock Learning to learn sequential network attacks using hidden markov models.
\newblock {\em IEEE Access}, 8:134480--134497.

\bibitem[Chatterjee and Namin, 2019]{chatterjee2019detecting}
Chatterjee, M. and Namin, A.-S. (2019).
\newblock Detecting phishing websites through deep reinforcement learning.
\newblock 2:227--232.

\bibitem[Chayal and Patel, 2021]{chayal2021review}
Chayal, N.~M. and Patel, N.~P. (2021).
\newblock Review of machine learning and data mining methods to predict different cyberattacks.
\newblock {\em Data Science and Intelligent Applications: Proceedings of ICDSIA 2020}, pages 43--51.

\bibitem[Chen et~al., 2019a]{chen2019applying}
Chen, C.-M., Wang, S.-H., Wen, D.-W., Lai, G.-H., and Sun, M.-K. (2019a).
\newblock Applying convolutional neural network for malware detection.
\newblock pages 1--5.

\bibitem[Chen et~al., 2019b]{chen2019automated}
Chen, Q., Islam, S.~R., Haswell, H., and Bridges, R.~A. (2019b).
\newblock Automated ransomware behavior analysis: Pattern extraction and early detection.
\newblock pages 199--214.

\bibitem[Chen et~al., 2021]{chen2021ai}
Chen, Y.-C., Chen, J.-L., and Ma, Y.-W. (2021).
\newblock Ai@ tss-intelligent technical support scam detection system.
\newblock {\em Journal of Information Security and Applications}, 61:102921.

\bibitem[Custers, 2021]{custers2021profiling}
Custers, B. (2021).
\newblock Profiling and predictions: challenges in cybercrime research datafication.
\newblock {\em Researching Cybercrimes: Methodologies, Ethics, and Critical Approaches}, pages 63--79.

\bibitem[Dasgupta et~al., 2022]{dasgupta2022machine}
Dasgupta, D., Akhtar, Z., and Sen, S. (2022).
\newblock Machine learning in cybersecurity: a comprehensive survey.
\newblock {\em The Journal of Defense Modeling and Simulation}, 19(1):57--106.

\bibitem[D{\.I}LEK et~al., 2015]{dilek2015applications}
D{\.I}LEK, S., {\c{C}}AKIR, H., and Ayd{\i}n, M. (2015).
\newblock Applications of artificial intelligence techniques to combating cyber crimes: A review.
\newblock {\em International Journal of Artificial Intelligence Applications (IJAIA)}, 6(1).

\bibitem[Egozi and Verma, 2018]{egozi2018phishing}
Egozi, G. and Verma, R. (2018).
\newblock Phishing email detection using robust nlp techniques.
\newblock pages 7--12.

\bibitem[Furnell and Dowling, 2019]{furnell2019cyber}
Furnell, S. and Dowling, S. (2019).
\newblock Cyber crime: a portrait of the landscape.
\newblock {\em Journal of Criminological Research, Policy and Practice}.

\bibitem[Gann, 2020]{gann2020hidden}
Gann, T. (2020).
\newblock The hidden costs of cybercrime on government.
\newblock \url{https://tinyurl.com/4eum556a}.

\bibitem[Gautam and Bansal, 2022a]{gautam2022effect}
Gautam, A.~K. and Bansal, A. (2022a).
\newblock Effect of features extraction techniques on cyberstalking detection using machine learning framework.
\newblock {\em Journal of Advances in Information Technology}, 13(5).

\bibitem[Gautam and Bansal, 2022b]{gautam2022performance}
Gautam, A.~K. and Bansal, A. (2022b).
\newblock Performance analysis of supervised machine learning techniques for cyberstalking detection in social media.
\newblock {\em Journal of Theoretical and Applied Information Technology}, 100(2):449--461.

\bibitem[Gautam and Bansal, 2023a]{gautam2023automatic}
Gautam, A.~K. and Bansal, A. (2023a).
\newblock Automatic cyberstalking detection on twitter in real-time using hybrid approach.
\newblock {\em International Journal of Modern Education and Computer Science}, 15(1):58.

\bibitem[Gautam and Bansal, 2023b]{gautam2023email}
Gautam, A.~K. and Bansal, A. (2023b).
\newblock Email-based cyberstalking detection on textual data using multi-model soft voting technique of machine learning approach.
\newblock {\em Journal of Computer Information Systems}, pages 1--20.

\bibitem[Go et~al., 2020]{go2020visualization}
Go, J.~H., Jan, T., Mohanty, M., Patel, O.~P., Puthal, D., and Prasad, M. (2020).
\newblock Visualization approach for malware classification with resnext.
\newblock pages 1--7.

\bibitem[Gogoi and Ahmed, 2022]{gogoi2022phishing}
Gogoi, B. and Ahmed, T. (2022).
\newblock Phishing and fraudulent email detection through transfer learning using pretrained transformer models.
\newblock pages 1--6.

\bibitem[Griffioen et~al., 2021]{griffioen2021scan}
Griffioen, H., Oosthoek, K., van~der Knaap, P., and Doerr, C. (2021).
\newblock Scan, test, execute: Adversarial tactics in amplification ddos attacks.
\newblock pages 940--954.

\bibitem[Guo et~al., 2022]{guo2022gld}
Guo, W., Qiu, H., Liu, Z., Zhu, J., and Wang, Q. (2022).
\newblock Gld-net: Deep learning to detect ddos attack via topological and traffic feature fusion.
\newblock {\em Computational Intelligence and Neuroscience}, 2022.

\bibitem[Han et~al., 2019]{han2019recognizing}
Han, X., Wang, L., Xu, S., Zhao, D., and Liu, G. (2019).
\newblock Recognizing roles of online illegal gambling participants: An ensemble learning approach.
\newblock {\em Computers \& Security}, 87:101588.

\bibitem[Holt and Lavorgna, 2021]{holt2021researching}
Holt, T.~J. and Lavorgna, A. (2021).
\newblock {\em Researching Cybercrimes: Methodologies, Ethics, and Critical Approaches}.
\newblock Springer.

\bibitem[Hou et~al., 2022]{hou2022identification}
Hou, Y., Wang, H., and Wang, H. (2022).
\newblock Identification of chinese dark jargons in telegram underground markets using context-oriented and linguistic features.
\newblock {\em Information Processing \& Management}, 59(5):103033.

\bibitem[Hughes et~al., 2021]{hughes2021too}
Hughes, J., Chua, Y.~T., and Hutchings, A. (2021).
\newblock Too much data? opportunities and challenges of large datasets and cybercrime.
\newblock {\em Researching Cybercrimes: Methodologies, Ethics, and Critical Approaches}, pages 191--212.

\bibitem[Hwang et~al., 2020]{hwang2020two}
Hwang, J., Kim, J., Lee, S., and Kim, K. (2020).
\newblock Two-stage ransomware detection using dynamic analysis and machine learning techniques.
\newblock {\em Wireless Personal Communications}, 112:2597--2609.

\bibitem[IBM, 2022]{ibm2022cost}
IBM (2022).
\newblock 2022 cost of a data breach report.
\newblock \url{https://www.ibm.com/resources/cost-data-breach-report-2022}.

\bibitem[Jha et~al., 2020]{jha2020recurrent}
Jha, S., Prashar, D., Long, H.~V., and Taniar, D. (2020).
\newblock Recurrent neural network for detecting malware.
\newblock {\em computers \& security}, 99:102037.

\bibitem[Khan et~al., 2020]{khan2020digital}
Khan, F., Ncube, C., Ramasamy, L.~K., Kadry, S., and Nam, Y. (2020).
\newblock A digital dna sequencing engine for ransomware detection using machine learning.
\newblock {\em IEEE Access}, 8:119710--119719.

\bibitem[Klein et~al., 2021]{klein2021plusmine}
Klein, J., Bhulai, S., Hoogendoorn, M., and Van~der Mei, R. (2021).
\newblock Plusmine: Dynamic active learning with semi-supervised learning for automatic classification.
\newblock pages 146--153.

\bibitem[Klein et~al., 2022]{klein2022jasmine}
Klein, J., Bhulai, S., Hoogendoorn, M., and van~der Mei, R. (2022).
\newblock Jasmine: A new active learning approach to combat cybercrime.
\newblock {\em Machine Learning with Applications}, 9:100351.

\bibitem[Kumar et~al., 2021]{kumar2021mcft}
Kumar, S. et~al. (2021).
\newblock Mcft-cnn: Malware classification with fine-tune convolution neural networks using traditional and transfer learning in internet of things.
\newblock {\em Future Generation Computer Systems}, 125:334--351.

\bibitem[Kumar and Janet, 2022]{kumar2022dtmic}
Kumar, S. and Janet, B. (2022).
\newblock Dtmic: Deep transfer learning for malware image classification.
\newblock {\em Journal of Information Security and Applications}, 64:103063.

\bibitem[Kumari et~al., 2018]{kumari2018machine}
Kumari, S., Saquib, Z., and Pawar, S. (2018).
\newblock Machine learning approach for text classification in cybercrime.
\newblock pages 1--6.

\bibitem[Maleh et~al., 2020]{maleh2020blockchain}
Maleh, Y., Shojafar, M., Alazab, M., and Romdhani, I. (2020).
\newblock Blockchain for cybersecurity and privacy: architectures, challenges, and applications.

\bibitem[Mittal et~al., 2022]{mittal2022deep}
Mittal, M., Kumar, K., and Behal, S. (2022).
\newblock Deep learning approaches for detecting ddos attacks: A systematic review.
\newblock {\em Soft Computing}, pages 1--37.

\bibitem[Moher et~al., 2009]{moher2009preferred}
Moher, D., Liberati, A., Tetzlaff, J., Altman, D.~G., and Group*, P. (2009).
\newblock Preferred reporting items for systematic reviews and meta-analyses: the prisma statement.
\newblock {\em Annals of internal medicine}, 151(4):264--269.

\bibitem[Morgan, 2022]{morgan2022top5}
Morgan, S. (2022).
\newblock Top 10 cybersecurity predictions and statistics for 2023.
\newblock \url{https://cybersecurityventures.com/stats/}.

\bibitem[Mos and Chowdhury, 2020]{mos2020growing}
Mos, M.~A. and Chowdhury, M.~M. (2020).
\newblock The growing influence of ransomware.
\newblock pages 643--647.

\bibitem[Mridha et~al., 2021]{mridha2021phishing}
Mridha, K., Hasan, J., Saravanan, D., and Ghosh, A. (2021).
\newblock Phishing url classification analysis using ann algorithm.
\newblock pages 1--7.

\bibitem[Mvula et~al., 2022]{mvula2022covid}
Mvula, P.~K., Branco, P., Jourdan, G.-V., and Viktor, H.~L. (2022).
\newblock Covid-19 malicious domain names classification.
\newblock {\em Expert Systems with Applications}, 204:117553.

\bibitem[Nahmias et~al., 2019]{nahmias2019trustsign}
Nahmias, D., Cohen, A., Nissim, N., and Elovici, Y. (2019).
\newblock Trustsign: trusted malware signature generation in private clouds using deep feature transfer learning.
\newblock pages 1--8.

\bibitem[Ngejane et~al., 2021]{ngejane2021digital}
Ngejane, C.~H., Eloff, J.~H., Sefara, T.~J., and Marivate, V.~N. (2021).
\newblock Digital forensics supported by machine learning for the detection of online sexual predatory chats.
\newblock {\em Forensic science international: Digital investigation}, 36:301109.

\bibitem[Oh et~al., 2020]{oh2020wipimization}
Oh, D.~B., Park, K.~H., and Kim, H.~K. (2020).
\newblock De-wipimization: Detection of data wiping traces for investigating ntfs file system.
\newblock {\em Computers \& Security}, 99:102034.

\bibitem[Palad et~al., 2019]{palad2019document}
Palad, E. B.~B., Tangkeko, M.~S., Magpantay, L. A.~K., and Sipin, G.~L. (2019).
\newblock Document classification of filipino online scam incident text using data mining techniques.
\newblock pages 232--237.

\bibitem[Pan and Yang, 2018]{pan2018cybersecurity}
Pan, J. and Yang, Z. (2018).
\newblock Cybersecurity challenges and opportunities in the new" edge computing+ iot" world.
\newblock pages 29--32.

\bibitem[Perry, 2013]{perry2013predictive}
Perry, W.~L. (2013).
\newblock {\em Predictive policing: The role of crime forecasting in law enforcement operations}.
\newblock Rand Corporation.

\bibitem[Phoka and Suthaphan, 2019]{phoka2019image}
Phoka, T. and Suthaphan, P. (2019).
\newblock Image based phishing detection using transfer learning.
\newblock pages 232--237.

\bibitem[Pradeepa and Devi, 2022]{pradeepa2022malicious}
Pradeepa, G. and Devi, R. (2022).
\newblock Malicious domain detection using nlp methods—a review.
\newblock pages 1584--1588.

\bibitem[Ravi et~al., 2021]{ravi2021adversarial}
Ravi, V., Alazab, M., Srinivasan, S., Arunachalam, A., and Soman, K. (2021).
\newblock Adversarial defense: Dga-based botnets and dns homographs detection through integrated deep learning.
\newblock {\em IEEE transactions on engineering management}, 70(1):249--266.

\bibitem[Ravi et~al., 2022a]{ravi2022recurrent}
Ravi, V., Chaganti, R., and Alazab, M. (2022a).
\newblock Recurrent deep learning-based feature fusion ensemble meta-classifier approach for intelligent network intrusion detection system.
\newblock {\em Computers and Electrical Engineering}, 102:108156.

\bibitem[Ravi et~al., 2022b]{ravi2022attention}
Ravi, V., Pham, T.~D., and Alazab, M. (2022b).
\newblock Attention-based multidimensional deep learning approach for cross-architecture iomt malware detection and classification in healthcare cyber-physical systems.
\newblock {\em IEEE Transactions on Computational Social Systems}.

\bibitem[Resende and Drummond, 2018]{resende2018survey}
Resende, P. A.~A. and Drummond, A.~C. (2018).
\newblock A survey of random forest based methods for intrusion detection systems.
\newblock {\em ACM Computing Surveys (CSUR)}, 51(3):1--36.

\bibitem[Reshmi, 2021]{reshmi2021information}
Reshmi, T. (2021).
\newblock Information security breaches due to ransomware attacks-a systematic literature review.
\newblock {\em International Journal of Information Management Data Insights}, 1(2):100013.

\bibitem[Rustam et~al., 2023]{rustam2023malware}
Rustam, F., Ashraf, I., Jurcut, A.~D., Bashir, A.~K., and Zikria, Y.~B. (2023).
\newblock Malware detection using image representation of malware data and transfer learning.
\newblock {\em Journal of Parallel and Distributed Computing}, 172:32--50.

\bibitem[Saad et~al., 2019]{saad2019curious}
Saad, S., Briguglio, W., and Elmiligi, H. (2019).
\newblock The curious case of machine learning in malware detection.
\newblock {\em Machine Learning Interpretability in Malware Detection}, 5:11.

\bibitem[Sachdeva and Ali, 2022]{sachdeva2022machine}
Sachdeva, S. and Ali, A. (2022).
\newblock Machine learning with digital forensics for attack classification in cloud network environment.
\newblock {\em International Journal of System Assurance Engineering and Management}, 13(Suppl 1):156--165.

\bibitem[Saha et~al., 2020]{saha2020phishing}
Saha, I., Sarma, D., Chakma, R.~J., Alam, M.~N., Sultana, A., and Hossain, S. (2020).
\newblock Phishing attacks detection using deep learning approach.
\newblock pages 1180--1185.

\bibitem[Salahdine and Kaabouch, 2019]{salahdine2019social}
Salahdine, F. and Kaabouch, N. (2019).
\newblock Social engineering attacks: A survey.
\newblock {\em Future Internet}, 11(4):89.

\bibitem[Salloum et~al., 2021]{salloum2021phishing}
Salloum, S., Gaber, T., Vadera, S., and Shaalan, K. (2021).
\newblock Phishing email detection using natural language processing techniques: a literature survey.
\newblock {\em Procedia Computer Science}, 189:19--28.

\bibitem[Sedjelmaci et~al., 2020]{sedjelmaci2020cyber}
Sedjelmaci, H., Guenab, F., Senouci, S.-M., Moustafa, H., Liu, J., and Han, S. (2020).
\newblock Cyber security based on artificial intelligence for cyber-physical systems.
\newblock {\em IEEE Network}, 34(3):6--7.

\bibitem[Shah et~al., 2019]{shah2019compromised}
Shah, S., Shah, B., Amin, A., Al-Obeidat, F., Chow, F., Moreira, F. J.~L., and Anwar, S. (2019).
\newblock Compromised user credentials detection in a digital enterprise using behavioral analytics.
\newblock {\em Future Generation Computer Systems}, 93:407--417.

\bibitem[Shams and Rizaner, 2018]{shams2018novel}
Shams, E.~A. and Rizaner, A. (2018).
\newblock A novel support vector machine based intrusion detection system for mobile ad hoc networks.
\newblock {\em Wireless Networks}, 24:1821--1829.

\bibitem[Sinaeepourfard et~al., 2019]{sinaeepourfard2019cybersecurity}
Sinaeepourfard, A., Sengupta, S., Krogstie, J., and Delgado, R.~R. (2019).
\newblock Cybersecurity in large-scale smart cities: novel proposals for anomaly detection from edge to cloud.
\newblock pages 130--135.

\bibitem[Singh et~al., 2021]{singh2021deep}
Singh, D., Shukla, A., and Sajwan, M. (2021).
\newblock Deep transfer learning framework for the identification of malicious activities to combat cyberattack.
\newblock {\em Future Generation Computer Systems}, 125:687--697.

\bibitem[Smith, 2020]{smith2020fbi}
Smith, R. (2020).
\newblock Fbi sees a 400$\%$ increase in reports of cyberattacks since the start of the pandemic.
\newblock \url{https://tinyurl.com/3xzvt8mr}.

\bibitem[Sun et~al., 2020]{sun2020deepdom}
Sun, X., Wang, Z., Yang, J., and Liu, X. (2020).
\newblock Deepdom: Malicious domain detection with scalable and heterogeneous graph convolutional networks.
\newblock {\em Computers \& Security}, 99:102057.

\bibitem[Vinayakumar et~al., 2019a]{vinayakumar2019deep}
Vinayakumar, R., Alazab, M., Soman, K., Poornachandran, P., Al-Nemrat, A., and Venkatraman, S. (2019a).
\newblock Deep learning approach for intelligent intrusion detection system.
\newblock {\em Ieee Access}, 7:41525--41550.

\bibitem[Vinayakumar et~al., 2019b]{vinayakumar2019robust}
Vinayakumar, R., Alazab, M., Soman, K., Poornachandran, P., and Venkatraman, S. (2019b).
\newblock Robust intelligent malware detection using deep learning.
\newblock {\em IEEE Access}, 7:46717--46738.

\bibitem[Wang et~al., 2023]{wang2023implementing}
Wang, L., Giang, C., Jerath, K., Raman, A., Lie, D., Chignell, M., et~al. (2023).
\newblock Implementing active learning in cybersecurity: Detecting anomalies in redacted emails.
\newblock {\em arXiv preprint arXiv:2303.00870}.

\bibitem[Wang et~al., 2022]{wang2022ssappidentify}
Wang, S., Yang, C., Guo, G., Chen, M., and Ma, J. (2022).
\newblock Ssappidentify: A robust system identifies application over shadowsocks’s traffic.
\newblock {\em Computer Networks}, 203:108659.

\bibitem[Wazirali et~al., 2021]{wazirali2021sustaining}
Wazirali, R., Ahmad, R., and Abu-Ein, A. A.-K. (2021).
\newblock Sustaining accurate detection of phishing urls using sdn and feature selection approaches.
\newblock {\em Computer Networks}, 201:108591.

\bibitem[Weiss and Khoshgoftaar, 2017]{weiss2017detection}
Weiss, K.~R. and Khoshgoftaar, T.~M. (2017).
\newblock Detection of phishing webpages using heterogeneous transfer learning.
\newblock pages 190--197.

\bibitem[Yadav et~al., 2022]{yadav2022efficientnet}
Yadav, P., Menon, N., Ravi, V., Vishvanathan, S., and Pham, T.~D. (2022).
\newblock Efficientnet convolutional neural networks-based android malware detection.
\newblock {\em Computers \& Security}, 115:102622.

\bibitem[Yin et~al., 2020]{yin2020apply}
Yin, J., Tang, M., Cao, J., and Wang, H. (2020).
\newblock Apply transfer learning to cybersecurity: Predicting exploitability of vulnerabilities by description.
\newblock {\em Knowledge-Based Systems}, 210:106529.

\bibitem[Yuan et~al., 2019]{yuan2019stealthy}
Yuan, K., Tang, D., Liao, X., Wang, X., Feng, X., Chen, Y., Sun, M., Lu, H., and Zhang, K. (2019).
\newblock Stealthy porn: Understanding real-world adversarial images for illicit online promotion.
\newblock pages 952--966.

\bibitem[Yusof et~al., 2019]{yusof2019systematic}
Yusof, A.~R., Udzir, N.~I., and Selamat, A. (2019).
\newblock Systematic literature review and taxonomy for ddos attack detection and prediction.
\newblock {\em International Journal of Digital Enterprise Technology}, 1(3):292--315.

\bibitem[Zaman et~al., 2022]{zaman2022trustworthy}
Zaman, S., Iqbal, M.~M., Tauqeer, H., Shahzad, M., and Akbar, G. (2022).
\newblock Trustworthy communication channel for the iot sensor nodes using reinforcement learning.
\newblock pages 1--6.

\bibitem[Zhao et~al., 2019]{zhao2019transfer}
Zhao, J., Shetty, S., Pan, J.~W., Kamhoua, C., and Kwiat, K. (2019).
\newblock Transfer learning for detecting unknown network attacks.
\newblock {\em EURASIP Journal on Information Security}, 2019:1--13.

\end{thebibliography}
\bibliographystyle{apalike}
\end{document}